\documentclass[10pt,twocolumn,letterpaper]{article}

\usepackage[pagenumbers]{cvpr} % To force page numbers, e.g. for an arXiv version

\usepackage[dvipsnames]{xcolor}
\newcommand{\red}[1]{{\color{red}#1}}
\newcommand{\blue}[1]{{\color{blue}#1}}

\definecolor{cvprblue}{rgb}{0.21,0.49,0.74}
\usepackage[pagebackref,breaklinks,colorlinks,citecolor=cvprblue]{hyperref}
\usepackage{listings}

\def\paperID{*****} % *** Enter the Paper ID here
\def\confName{CVPR}
\def\confYear{2024}

\title{Solving Situation Puzzles with Large Language Model and External Reformulation}

\author{
\begin{tabular}{@{}c@{\hspace{1.2cm}}c@{\hspace{1.2cm}}c@{\hspace{1.2cm}}c@{}}
{\small Kun Li} & {\small Xinwei Chen} & {\small Tianyou Song} & {\small Chengrui Zhou} \\
{\footnotesize University of Illinois Urbana-} & {\footnotesize University of Illinois at Urbana} & {\footnotesize Columbia University} & {\footnotesize Columbia University} \\
{\footnotesize Champaign} & {\footnotesize Champaign} & {\footnotesize New York City, NY, USA} & {\footnotesize New York City, USA} \\
{\footnotesize Champaign, IL, USA} & {\footnotesize Champaign, IL, USA} & {\footnotesize tianyou.song@columbia.edu} & {\footnotesize zhou.chengrui@columbia.edu} \\
{\footnotesize kunli3@illinois.edu} & {\footnotesize xinweic2@illinois.edu} & & \\[1em]
{\small Zhuoran Liu} & {\small Zhenyan Zhang} & {\small Jiangjian Guo} & {\small Qing Shan*} \\
{\footnotesize Carnegie Mellon University} & {\footnotesize Carnegie Mellon University} & {\footnotesize University of California San} & {\footnotesize Northeastern University} \\
{\footnotesize Pittsburgh, USA} & {\footnotesize Pittsburgh, USA} & {\footnotesize Diego} & {\footnotesize Seattle, WA, USA} \\
{\footnotesize zliu3@alumni.cmu.edu} & {\footnotesize zhenyan2@alumni.cmu.edu} & {\footnotesize La Jolla, CA, USA} & {\footnotesize shan.qi@northeastern.edu} \\
& & {\footnotesize j9guo@ucsd.edu} & \\
\end{tabular}
}

\begin{document}
\maketitle

\begin{abstract}
    In recent years, large language models (LLMs) have shown an impressive ability to perform arithmetic and symbolic reasoning tasks. However, we found that LLMs (\eg, ChatGPT) cannot perform well on reasoning that requires multiple rounds of dialogue, especially when solving situation puzzles. Specifically, LLMs intend to ask very detailed questions focusing on a specific aspect or same/similar questions after several rounds of Q\&As. To help LLMs get out of the above dilemma, we propose to integrate LLMs with external reformulation, where the situation puzzle will be reformulated after several rounds of Q\&A or when the LLMs raise an incorrect guess. Experiments show superior performance (\eg, win rate, number of question/guess attempts) of our method than directly using LLMs for solving situation puzzles.
\end{abstract}
\section{Introuction}

Natural language processing (NLP) has made significant progress in recent years, especially with the introduction of transformers and pre-trained language models. However, their ability to perform natural language reasoning is still far from satisfactory. NLR, the process of reasoning based on existing knowledge, is a fundamental aspect of human intelligence and is critical for complex tasks such as comprehension of complex and abstract situations, and decision-making. Building artificial intelligence systems with reasoning capabilities is not only the ultimate goal of the research but also a necessary way to improve the performance of complex applications.

In recent years, advancements in NLP have led to the development of increasingly sophisticated LLMs (large language models)~\cite{radford2018improving}, capable of performing a wide range of language-based tasks with remarkable proficiency. Among these models, ChatGPT, developed by OpenAI, has emerged as a frontrunner, showcasing abilities that often parallel human-like language understanding and generation~\cite{lock2022ai}. However, despite these advancements, LLMs continue to exhibit limitations in certain complex cognitive tasks, particularly those requiring deep reasoning and understanding beyond surface-level information processing. One such challenging domain is the realm of situation puzzles, which demand a high degree of creative and lateral thinking.

\begin{figure}[t]
\centering
\includegraphics[width=1.0\linewidth]{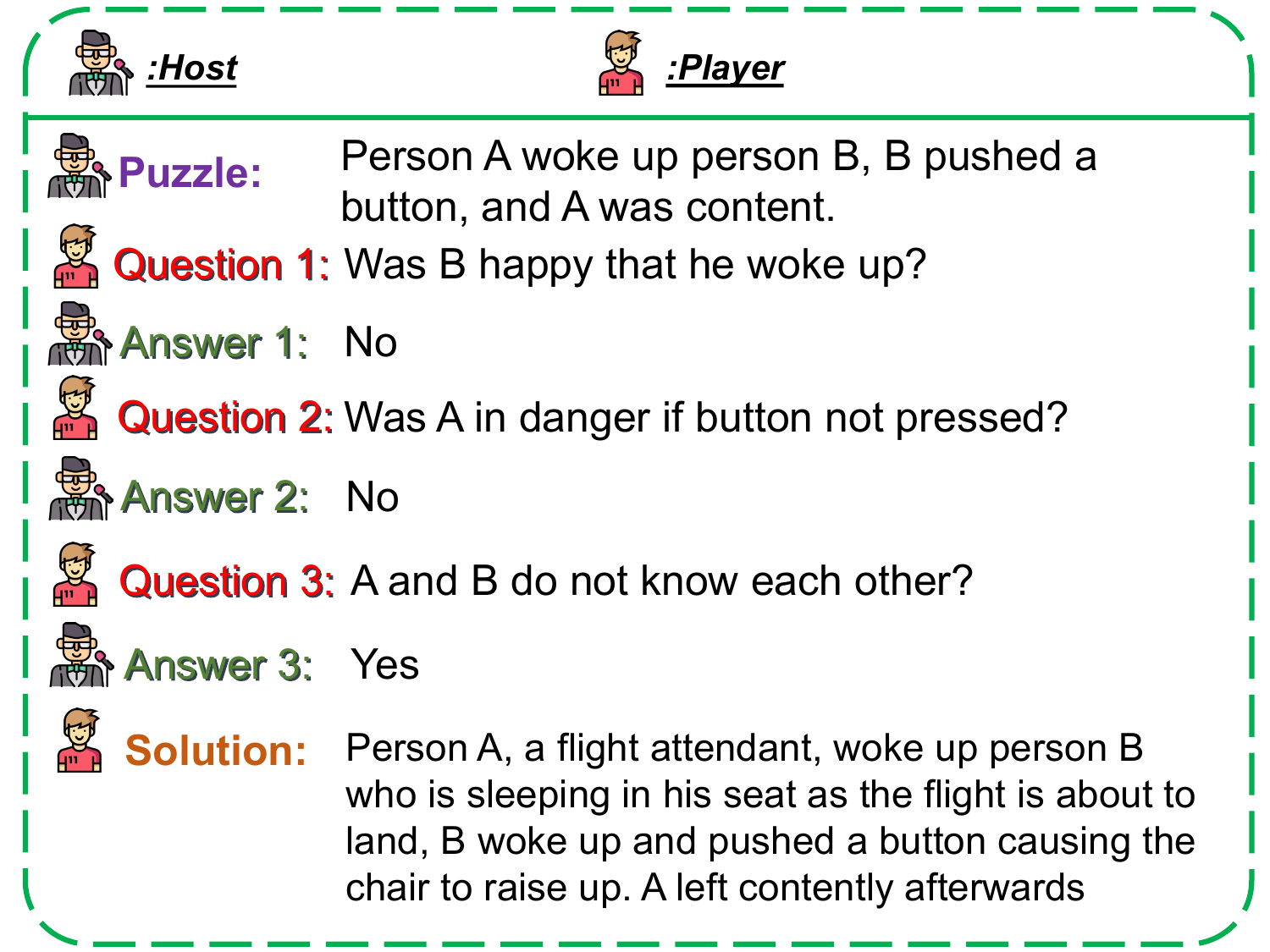}
\vspace{-7.5mm}
\caption{An example of solving a situation puzzle, including several rounds of interaction between the host and the player.}
\label{fig:example_suitation_puzzle}
\end{figure}

\begin{figure*}[t]
\centering
\includegraphics[width=1.0\linewidth]{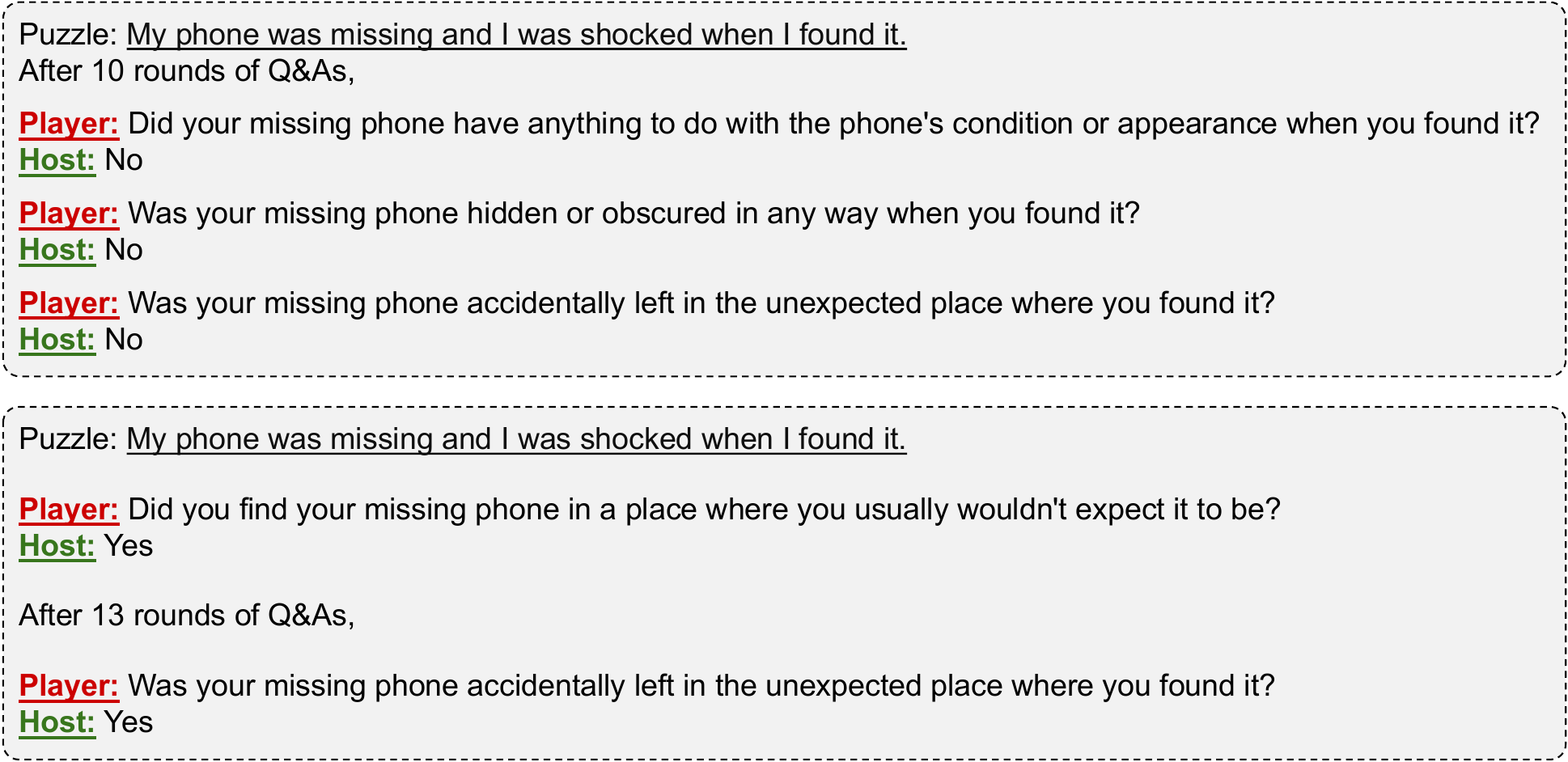}
\vspace{-6mm}
\caption{Two examples of directly using LLMs to solve situation puzzles. For the example in the top, after several rounds of Q\&As, the player intends to ask very detailed questions focusing on a specific aspect. For the example at the bottom, the player asks the same or similar questions after several rounds of Q\&As.}
\label{fig:baseline_example}
\end{figure*}

Situation puzzles, often referred to as lateral thinking puzzles, are a unique and intellectually stimulating class of riddles. These puzzles present scenarios with limited information, requiring solvers to ask targeted questions and make logical inferences to uncover the underlying story or solution, as the example given in Fig.~\ref{fig:example_suitation_puzzle}. The ability to solve such puzzles is indicative of a sophisticated level of cognitive processing, blending creativity, reasoning, and inference - skills that are inherently human and have been historically challenging for AI to replicate. When applying LLMs to solve situation puzzles, we observe the player (LLMs) often get stuck in a long dialog. Specifically, as examples shown in Fig.~\ref{fig:baseline_example}, after multiple ($\sim$10) rounds of Q\&As, the player intends to ask very detailed questions or similar questions. These behaviors indeed contribute trivially to solving the puzzle. 

\begin{figure}[t]
\centering
\includegraphics[width=1.0\linewidth]{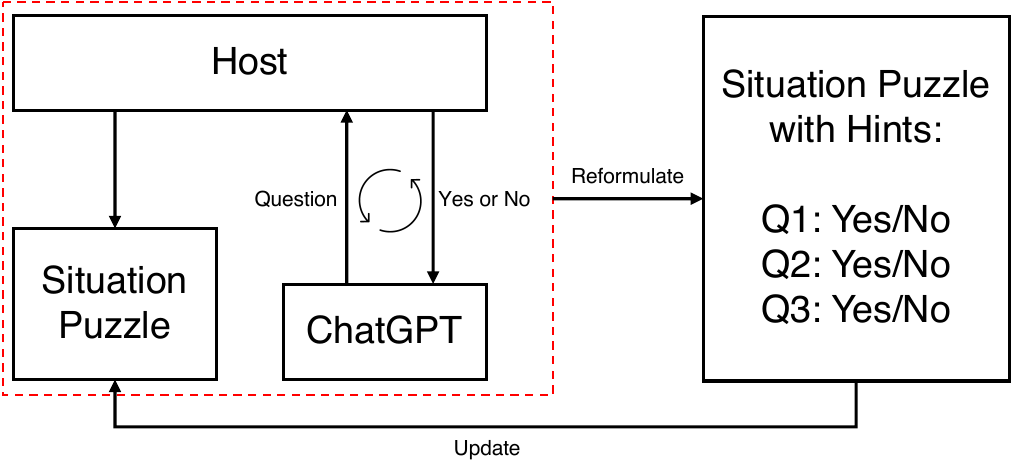}
\vspace{-8mm}
\caption{The key idea of our proposed reformulation manner for LLMs to solve the situation puzzles.}
\label{fig:overview}
\end{figure}

To address the above limitations and help LLMs get out of the long-dialog dilemma, we propose to reformulate the situation description with previously asked Q\&As by the LLMs and then restart a new chat session to solve the puzzle. As the overview shown in Fig.~\ref{fig:overview}, to reformulate, previous Q\&As are selected as additional hints to formulate the situation description. And the new chat session will start with the reformulated description. Note that the puzzle can be iteratively reformulated once the reformulation conditions are satisfied. By incorporating reformulation, LLMs can easily get out of the above dilemma and quickly come out with the correct answer. To validate, we have conducted experiments to compare our reformulation method with directly using LLMs to solve situation puzzles, which demonstrate considerable improvements.

\section{Related Work}
\subsection{LLMs for Problem Solving}
Currently, with the introduction of ChatGPT and GPT-4 (OpenAI, 2023), Large Language Models (LLMs) are gaining increasingly potent capabilities, which enable them to tackle a broad range of tasks effectively. Many benchmarks and research efforts have been dedicated to evaluating LLMs from various perspectives in order to discover whether LLMs think as humans or just merely retrieve from their vast world knowledge, including language tasks~\cite{liang2022holistic, bosselut2022proceedings, moens2021findings}, reasoning~\cite{bang2023multitask, bian2023chatgpt}, robustness~\cite{li2023survey}, trustworthiness~\cite{hagendorff2023human}, medical applications~\cite{chervenak2023promise,cascella2023evaluating}, and ethical considerations~\cite{cao2023assessing,parrish2021bbq}. However, those previous tasks do not require multi-turn interactions to thoroughly assess the interactive abilities of LLMs and also lack standardized evaluation criteria for measuring the divergent thinking capabilities of LLMs. To tackle this problem, some research focuses on planning and reasoning~\cite{valmeekam2022large}, question answering~\cite{hao2022recent}, lateral thinking puzzle~\cite{huang2023lateval,jiang2023brainteaser} enabling the evaluation of the divergent thinking capabilities of LLMs. show that the LLMs lack of ability to employ divergent thinking ability during interactions.

\subsection{Situation Puzzle}
The situation puzzle, also called as Lateral Thinking Puzzle, is one of the most popular among enthusiasts of inference, focusing on finding the single best solution to a problem, involving analyzing the problem step-by-step and narrowing down possibilities until the optimal solution is reached. \cite{syahrin2019creative} consider that lateral thinking is a cognitive activity employed to construct creative ideas. For the generation of novel ideas that promote progress in different scientific fields from engineering to the arts, politics to personal well-being, research~\cite{hidayat2018effectiveness} denotes that lateral thinking is oriented to creative thinking skills and problem-solving. Relaiza et al.~\cite{relaiza2021cognitive} indicate that lateral thinking can offer alternative approaches and perspectives to vertical thinking, thereby augmenting its efficacy. Research~\cite{valmeekam2022large,hao2022recent,huang2023lateval,jiang2023brainteaser} argue that the LLMs have shown powerful performance in vertical thinking, however, lack the ability in lateral thinking. Hence, to build an external tool and reformulate situation puzzles to help LLMs rethink the question laterally.

\subsection{Reformulation of Prompts for LLMs}
The LLMs show their strong ability to solve different kinds of problems compared to conventional models~\cite{liang2022holistic, bosselut2022proceedings, moens2021findings,bang2023multitask, bian2023chatgpt,li2023survey,hagendorff2023human,chervenak2023promise,cascella2023evaluating,cao2023assessing,parrish2021bbq}. However, for complex questions~\cite{hao2022recent,valmeekam2022large,huang2023lateval,jiang2023brainteaser}, the LLMs are not able to give a satisfactory answer. Especially, there exists a strong
need for improving the systematic and extensible planning ability of LLMs and bestowing LLMs with innate planning capabilities. Some pioneer research, such as~\cite{valmeekam2022large,hao2022recent,huang2023lateval,jiang2023brainteaser,juneja2023small,juneja2023small,valmeekam2023planbench}, propose the reformulation method to solve the complex problem based on LLMs. \cite{juneja2023small} proposed to use the small language model fine-tuned to coordinate the large language model to enhance the ability of reasoning capabilities of LLMs. The PlanBanch~\cite{valmeekam2023planbench} proposed different reformulation methods for LLMs in planning or reasoning about actions and change. \cite{mishra2021reframing} proposed the reframing instructional prompts to reformulate a complex task into multiple subtasks by using multi-step prompting. In summary, most of the current research focuses on reformulating different tasks in a specified designed manner. Hence for situation puzzles, our approach follows the previous research and proposes an external reformulation method.
\section{Methods}

In this section, we formally formulate the problem of solving situation puzzles, and then describe our proposed reformulation manner. Finally, we explain how to implement the baseline method and conduct reformulation.

\subsection{Problem Formulation}

We refer to the game as solving a given situation puzzle. There are two roles in the game, \ie, the player and the host. At the beginning of the game, the host will give the description of the situation puzzle, then the player starts asking questions, and the host will answer each question with one of ``Yes'', ``No'', or ``Irrelevant''. Sometimes, the player will give a guess to the puzzle, and the answer from the host should be ``Yes'' or ``No''. The game will end if 1.) the player gives a correct guess, 2.) the number of guess attempts reaches the given maximum, or 3.) the number of question attempts reaches the given maximum. When condition 1.) satisfied, the player wins the game, otherwise, the player loses the game.

To formulate, we denote the description of the puzzle as $\mathcal{S}$, the answer to the puzzle as $\mathcal{A}$, the question as $q$, the answer to the question as $r \in \{\text{Yes}, \text{No}, \text{Irrelevant}\}$, the guess as $g$, and the answer to the guess as $e \in \{\text{Yes}, \text{No}\}$. Assume that the player has asked $i$ questions and given $k$ guesses. Then, in the next round of chatting, the player has two possible actions: asking a question or giving a guess.
If the player asks a question,
\begin{equation}
    q_{i+1} = \text{Player.ask}\Big(\mathcal{S}, \big\{(q_j, r_j)\big\}_{j=1}^{i}, \big\{(g_j, e_j)\big\}_{j=1}^{k}\Big),
\end{equation}
the host will give the answer $r_{i+1}$ to the question $q_{i+1}$, \ie, 
\begin{equation}
    r_{i+1} = \text{Host.answer}(\mathcal{S}, \mathcal{A}, q_{i+1}).
\end{equation}
If the player gives a guess,
\begin{equation}
    g_{k+1} = \text{Player.guess}\Big(\mathcal{S}, \big\{(q_j, r_j)\big\}_{j=1}^{i}, \big\{(g_j, e_j)\big\}_{j=1}^{k}\Big),
\end{equation}
the host will give the answer $e_{k+1}$ to the guess $g_{k+1}$, \ie, 
\begin{equation}
    e_{k+1} = \text{Host.answer}(\mathcal{S}, \mathcal{A}, g_{k+1}).
\end{equation}
Particularly, if $e_{k+1}$ is ``Yes'', the player wins the game and the game ends, otherwise, the game continues.

\begin{figure*}[t!]
\centering
\includegraphics[width=1.0\linewidth]{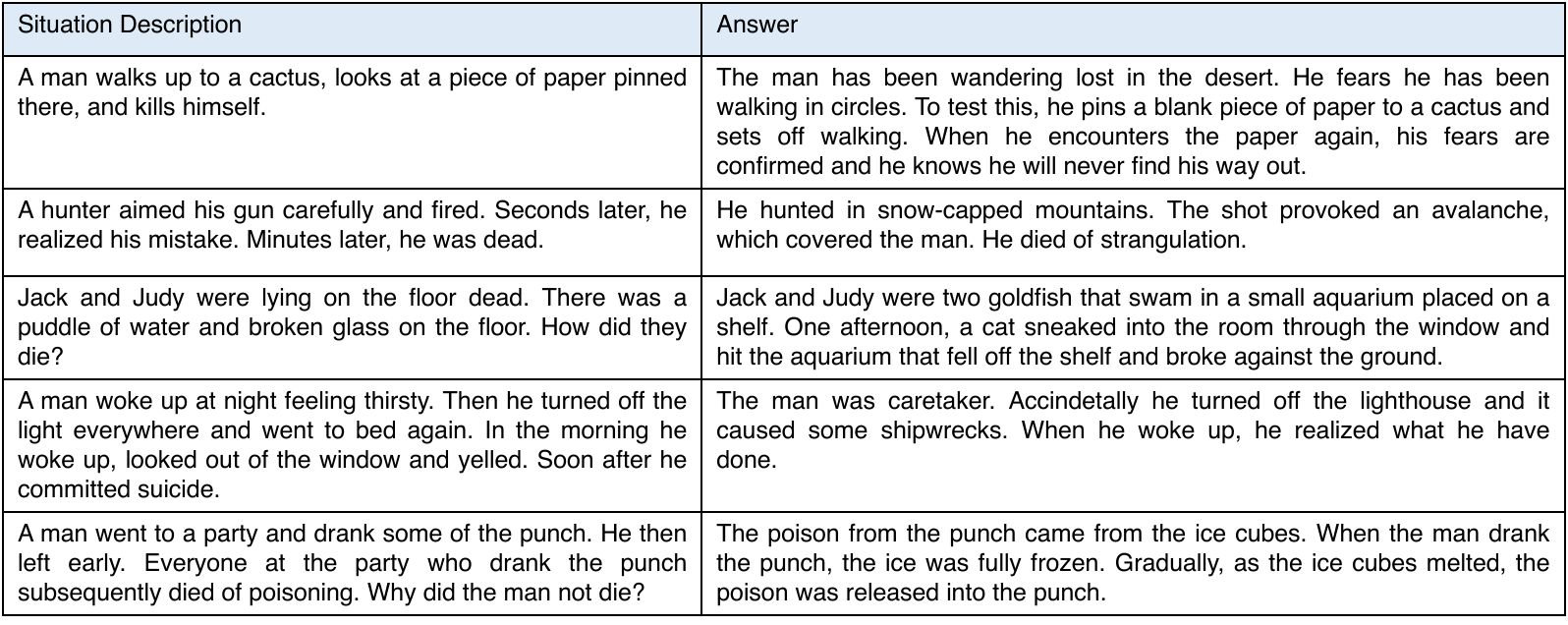}
\vspace{-7.5mm}
\caption{We conducted experiments on the above five situation puzzles.}
\label{fig:cases}
\end{figure*}

\subsection{External Reformulation}

We define the concept of the chat session. At the beginning of $c$-th chat session, the host gives the description of the situation puzzle $\mathcal{S}_c$, then the player asks questions/guesses, and the host answers them. Our idea is to end the chat session after several rounds of Q\&As, reformulate the description with Q\&As, and then start a new chat session. Here, we preset two conditions to ending the chat session: 1.) the player gives an incorrect guess (\ie, the answer to the guess is ``No''), 2.) the player has asked $K$ questions in this chat session. If one of the above conditions is satisfied, we end the chat session and start a new chat session with the reformulated description.

To formulate, assume at $c$-th chat session, the player asked $k_c (k_c \leq K)$ questions. We denote $k_c$ Q\&A pairs as $\mathcal{Q}_c = \{(q_j, r_j)\}_{j=1}^{k_c}$. Instead of selecting all Q\&As for reformulation, we define the priority of questions by their answers and select top-$M$ Q\&As pairs, where ``Yes'' is higher than ``No'' and ``No'' is higher than ``Irrelevant'' (\ie, ``Yes'' $>$ ``No'' $>$ ``Irrelevant''). In practice, all Yes-questions (the answer is ``Yes'') are selected as they are more related to the answer $\mathcal{A}$ of the puzzle. If Yes-questions are fewer than $M$, we select the rest from No/Irrelevant-questions (the answer is ``No''/``Irrelevant'') by their priorities (``No'' $>$ ``Irrelevant'').

Then, suppose that there are $M_c$ Q\&As are selected, denoted as $\mathcal{Q}_c' \subset \mathcal{Q}_c$. Note that $M_c$ can be either larger than $M$ (the player asked more than $M$ Yes-questions) or smaller than $M$ (the player gives an incorrect guess with fewer than $M$ questions being asked). Based on the above, we reformulate the description of the puzzle with $\mathcal{Q}_c'$,
\begin{equation}
    \mathcal{S}_{c+1} = \text{Reformulate}(\mathcal{S}_c, \mathcal{Q}_c').
\end{equation}
Then, a new chat session is started with $\mathcal{S}_{c+1}$.

\subsection{Implementation}

Assume that the original description of the puzzle is $\mathcal{S}_0$, the prompt is given as

\vspace{6pt}
\blue{
\textit{Solve the following situation puzzle and guess the reason. You can ask questions, and I will give the answer yes/no or irrelevant. Once you want to give a guess, please start with ``I guess that ...''}

\textit{Description: $\mathcal{S}_0$}
}
\vspace{6pt}

\noindent
For reformulation, we define the following prompt

\vspace{6pt}
\blue{
\textit{Solve the following situation puzzle and guess the reason. You can ask questions, and I will give the answer yes/no or irrelevant. Once you want to give a guess, please start with ``I guess that ...''}

\textit{Description: $\mathcal{S}_0$}
}

\red{
\textit{Here are some hints:}

\textit{1. ...}

\textit{2. ...}
}

\vspace{6pt}

\noindent
Each hint is generated from one pair of questions and answers. For example, for the question ``Was the man lost in a desert?'' with the answer of ``Yes'', the hint can be ``The man was lost in the desert.'' In practice, we utilize ChatGPT to automatically generate hints from selected Q\&As.

In the baseline, the game starts and ends in the same chat session. In our method, the game starts with the original description, and the chat session can be ended and restarted with reformulated descriptions. When the game ended, we recorded the results, including win/lose, the number of questions, the number of guess attempts, and the number of Yes/No/Irrevelant-questions as evaluation metrics for comparison.

\section{Experiments}

\begin{table}[t]
    \centering
    \caption{Comparison of baseline and our methods on solving the situation puzzle. Several evaluation metrics are compared including win/lose and the number of asked questions/guesses.}
    \vspace{-6pt}
    \setlength{\tabcolsep}{8pt}
    \begin{tabular}{l|c|c}
    \hline
        Metrics & Baseline & Ours \\ \hline \hline
        Win/Lose & 2/3 & 4/1 \\
        \# Guesses & 2.8 & 2.4 \\
        \# Questions & 28.6 & 22.6 \\
        \# Yes-Questions & 3.6 & 5 \\
        \# No-Questions & 18.2 & 12.2 \\
        \# Irrelevant-Questions & 6.8 & 5.4 \\
    \hline
    \end{tabular}
    \label{tab:main}
\end{table}

\begin{table*}[t]
    \centering
    \caption{Ablation study on different conditions to reformulate the situation description. ``Wrong Guess Only'' means that the chat session ends only when the player gives an incorrect guess.}
    \vspace{-6pt}
    \setlength{\tabcolsep}{8pt}
    \begin{tabular}{l|c|c|c}
    \hline
        Metrics & $K=5, M=3$ (best) & Wrong Guess Only & $K=10, M=6$ \\ \hline \hline
        Win/Lose & 4/1 & 3/2 & 3/2 \\
        \# Guesses & 2.4 & 4 & 3 \\
        \# Questions & 22.6 & 22.8 & 24.2 \\
        \# Yes-Questions & 5 & 6.8 & 6 \\
        \# No-Questions & 12.2 & 12 & 10.8 \\
        \# Irrelevant-Questions & 5.4 & 4 & 7.4 \\
    \hline
    \end{tabular}
    \label{tab:ablation}
\end{table*}

\begin{figure*}[t!]
\centering
\includegraphics[width=1.0\linewidth]{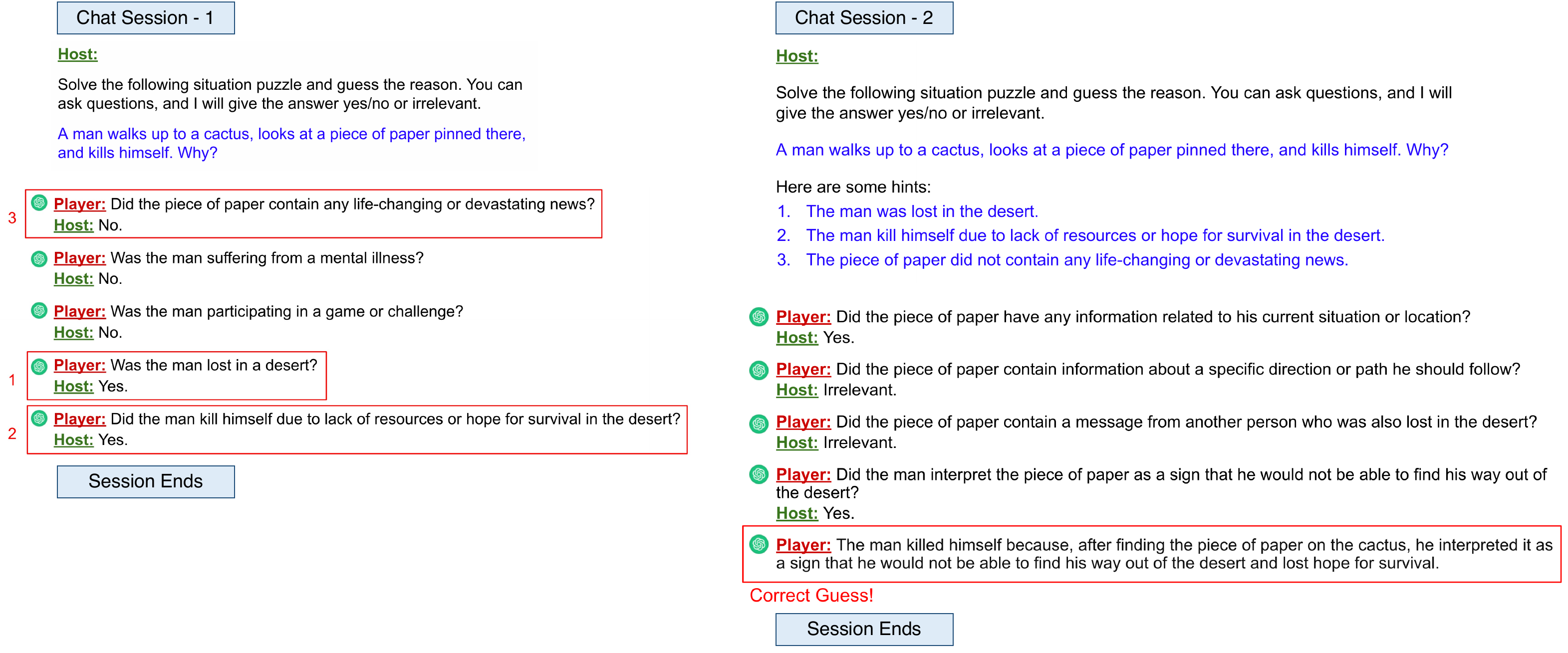}
\vspace{-6mm}
\caption{Case study. In the first chat session, the game starts with the host giving a description of the situation. After 5 rounds of Q\&As, two Yes-questions and the first No-question are selected to generate the hints. To reformulate, hints are integrated into the description prompt and a new chat session starts with the new description. In this case, the game ends in the second chat session as the player gives a correct guess and finally wins the game.}
\label{fig:example}
\end{figure*}

To validate the effectiveness of the proposed reformulation manner, we conducted experiments on 5 different situation puzzles given in Fig.~\ref{fig:cases}. We preset the maximum number of questions and guesses to 30 and 5, respectively. In our method, we choose $K=5$ and $M=3$. Additionally, we compare the performance of different selections of $K/M$ in the ablation study.

\vspace{6pt}
\noindent
\textbf{Results.}
Table~\ref{tab:main} compares the performance of the baseline model and our method of solving situation puzzles, where the number of game win/lose, questions, guesses, and Yes/No/Irrevelant-questions are recorded and compared. By incorporating the reformulation, the player has a higher win rate with fewer questions and guesses being asked. Additionally, the player asked more Yes-questions and fewer No/Irrelevant-questions, which means that reformulation indeed helped the player (LLMs) to get out of the dilemma and ask more questions related to the correct puzzle answer.

\vspace{6pt}
\noindent
\textbf{Ablation Study.}
In Table~\ref{tab:ablation}, we compare different conditions to reformulate the situation description. In addition to comparing different selections of $K/M$, we also tried to reformulate the description only when the player gives an incorrect guess (``Wrong Guess Only''). The experimental results show that the best selection of $K/M$ is 5/3, and it would be important to end the chat session early by setting the maximum number of questions being asked.

\vspace{6pt}
\noindent
\textbf{Case Study.}
In Fig.~\ref{fig:example}, we conduct a case study to show how the reformulation works to help the LLMs solve situation puzzles. First, we start a chat session, the host provides the original description, and the player asks questions until the one of reformulation conditions is satisfied. After 5 rounds of Q\&As, we end the chat session and select all two Yes-questions and the first No-question to reformulate the situation description. Then, a new chat session starts with the reformulated description, and the player asks questions and finally gives the correct guess.

% \vfill

% \section{Conclusion}

% In this project, we study the problem of solving situation puzzles with large language models (LLMs). We observed that LLMs often get stuck after multiple rounds of Q\&As. Specifically, the player (LLMs) intends to ask very detailed questions or the same/similar questions. To address this, we propose to reformulate the situation description with previous Q\&As and restart a new chat session to solve the new problem. Reformulation can help LLMs quickly get out of the dilemma of long dialog while previous Q\&As (experiences/knowledge) are still retained. Experiments are conducted to demonstrate the effectiveness of the reformulation manner.

\section{Conclusion}

This study investigates the problem-solving capabilities of large language models (LLMs) when applied to situation puzzles. Our observations reveal that LLMs frequently encounter performance plateaus after extended question-answer exchanges, manifesting as either excessively granular inquiries or redundant questioning patterns.

To address this limitation, we introduce a novel reformulation methodology wherein the original problem statement is augmented with accumulated Q\&A interactions, facilitating the initiation of a new dialogue session. This approach effectively circumvents the cognitive constraints associated with prolonged conversational exchanges while preserving the knowledge capital acquired during previous interactions. The reformulation technique enables LLMs to overcome conversational inertia by providing a consolidated knowledge foundation rather than requiring the maintenance of extensive dialogue history. This restructuring of the problem space creates a more efficient starting point that incorporates all salient discoveries from prior exchanges.

Through systematic empirical evaluation across diverse puzzle scenarios and model architectures, we demonstrate that our reformulation approach significantly outperforms traditional continuous dialogue methods in terms of both solution efficiency and question redundancy reduction. The experimental results validate the efficacy of our methodology and suggest its potential application for enhancing reasoning capabilities in complex, multi-step cognitive tasks.

These findings contribute to the growing body of research on improving LLM performance in tasks requiring sustained reasoning and strategic information gathering.

{
    \newpage
    \small
    \bibliographystyle{ieeenat_fullname}
    \bibliography{main}
}

\end{document}